\documentclass[journal]{IEEEtran}
\usepackage{graphicx}
\usepackage{graphics} 
\usepackage{epsfig} 
\usepackage{amssymb}  
\usepackage{graphics} 
\usepackage{bm,upgreek}
\usepackage{amsmath}
\usepackage{multirow}
\usepackage{float}
\usepackage[ruled,vlined]{algorithm2e}
\usepackage[compatible]{algpseudocode} 
\usepackage{cite}
\usepackage[dvipsnames]{xcolor}
\usepackage{tikz}
\usepackage{url}
\usepackage{cite}

\DeclareMathOperator*{\argmax}{arg\,max}

\makeatother

\begin{document}

\title{Multi-Goal Dexterous Hand Manipulation using Probabilistic Model-based Reinforcement Learning}

\author{Yingzhuo Jiang$^{1,2}$, Wenjun Huang$^{1,2}$, Rongdun Lin$^{1,2}$, Chenyang Miao$^{1,2}$, Tianfu Sun$^{2}$ and Yunduan Cui$^{2,*}$
\thanks{This research is supported by the National Natural Science Foundation of China under Grants 62103403.}
\thanks{$^{1}$ University of Chinese Academy of Sciences, Beijing, China}
\thanks{$^{2}$ Shenzhen Institutes of Advanced Technology, Chinese Academy of Sciences, Shenzhen, China.}
\thanks{$^*$ Corresponding author: Yunduan Cui (e-mail: cuiyunduan@gmail.com)}
}

\maketitle
\begin{abstract}
This paper tackles the challenge of learning multi-goal dexterous hand manipulation tasks using model-based Reinforcement Learning. We propose Goal-Conditioned Probabilistic Model Predictive Control (GC-PMPC) by designing probabilistic neural network ensembles to describe the high-dimensional dexterous hand dynamics and introducing an asynchronous MPC policy to meet the control frequency requirements in real-world dexterous hand systems.
Extensive evaluations on four simulated Shadow Hand manipulation scenarios with randomly generated goals demonstrate GC-PMPC's superior performance over state-of-the-art baselines. It successfully drives a cable-driven Dexterous hand, DexHand 021 with 12 Active DOFs and 5 tactile sensors, to learn manipulating a cubic die to three goal poses within approximately 80 minutes of interactions, demonstrating exceptional learning efficiency and control performance on a cost-effective dexterous hand platform.
\end{abstract}

\section{Introduction}

\IEEEPARstart{A}{chieving} human-like control of dexterous robotic hands remains a critical challenge in robotics. Traditional control methods are limited by the complexity of modeling hand dynamics and planning hand-object interactions. These limitations prevent the implementation of advanced manipulation tasks, undermining the key advantages of dexterous hands such as superior flexibility and adaptability compared to standard robotic grippers.
Reinforcement learning (RL), which autonomously discovers optimal control policies through exploration without prior knowledge of the target systems~\cite{kober2013reinforcement}, has emerged as a promising alternative to traditional control methods in dexterous hand manipulation.
Traditional RL based on kernel function approximation has been 
applied to control two fingers of a pneumatic Shadow Hand to learn a bottle-cap unscrewing task through discrete control actions~\cite{cui2017kernel}.
Later research investigated the impact of human demonstrations on deep RL on a simulated 24-DoF ADROIT hand~\cite{rajeswaran2018learning}.
With the guidance of human demonstrations, Model-free RL was successfully applied to Dclaw and Allegro hands to learn valve rotation tasks in specific scenarios~\cite{r3}.
Sim2Real technology also played a crucial role in RL applications on dexterous hands. 
Through extensive simulation training, it achieved superior generalization ability in vision-based object grasping task using low-cost dexterous hand hardware~\cite{r8}. It also transferred the RL policies learned in simulations to real-world dexterous hands, demonstrating effective manipulation of both concave elongated cubes~\cite{r6} and slender cylindrical objects~\cite{r5}.

Despite the progressively improving performance of RL in dexterous manipulation, state-of-the-art approaches still depend heavily on high-quality simulation and comprehensive sensor data from real hardware~\cite{r7}. Moreover, dexterous manipulation often involves multiple objectives, such as rotating cubes to specific poses or turning valves to precise angles, which can be categorized as multi-goal tasks. 
These tasks force traditional RL approaches to spend tens of hours learning the corresponding value functions while simultaneously struggling with policy instability due to the expansive task space and sparse reward distributions~\cite{ijcai2022p770}.

\begin{figure}[t]
\centering
\includegraphics[width=0.9\columnwidth]{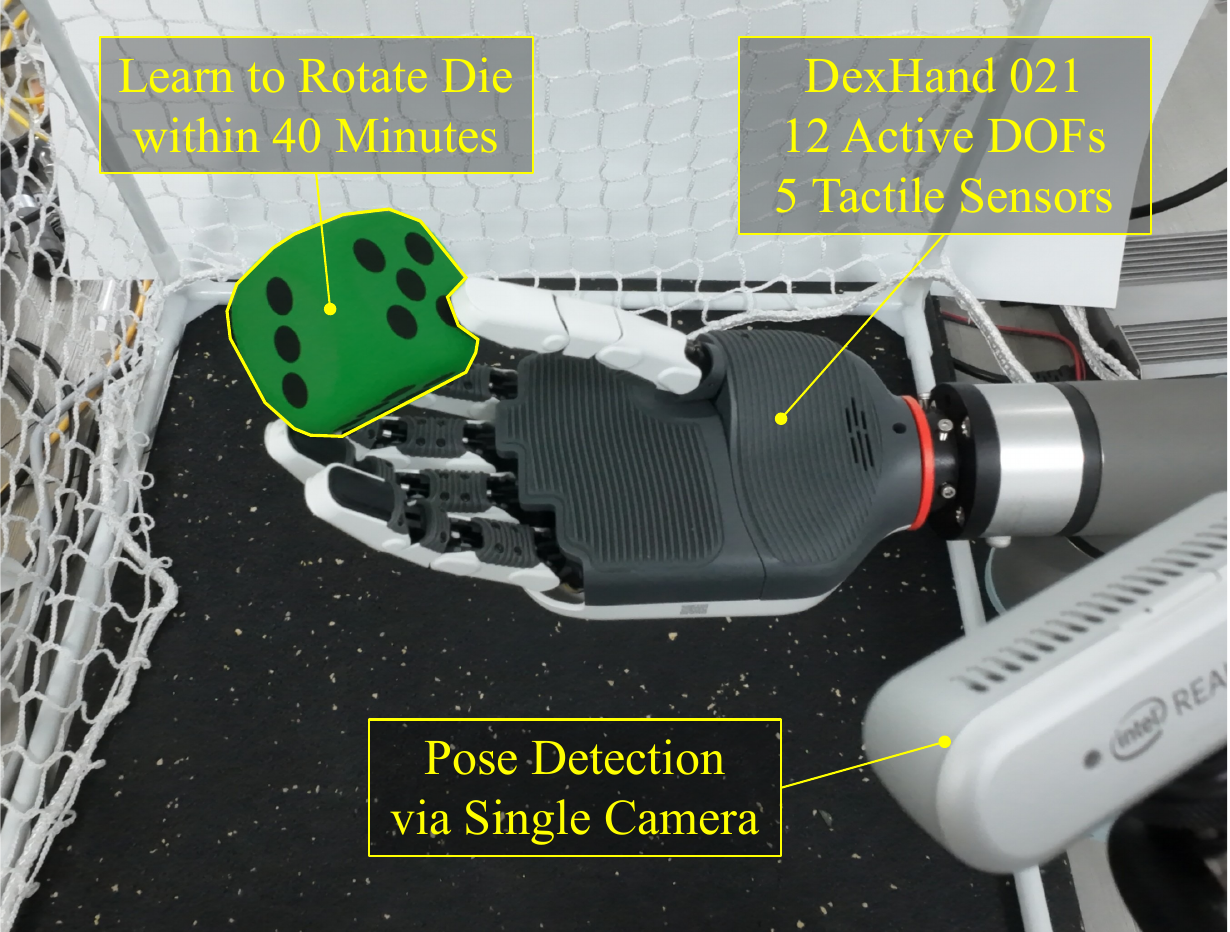}
\caption{Real-world dexterous hand used in this work. It was successfully drove to learn a multi-goal die rotation task within $80$ minutes based on a single-camera pose detection.}
\label{figure:1}
\end{figure}

Model-based RL (MBRL) attempts to model the dynamics of the target system during the interactions and leverages the learned model to effectively explore the optimal policies~\cite{r12}. It was considered as one of the promising solutions for solving multi-goal tasks~\cite{ijcai2021p480} because accurately estimating the target system's dynamics can avoid explicitly learning the relationship between states and varying task goals.
Google Brain developed a deep neural network model of Shadow hand dynamics and achieved Baoding balls manipulation by integrating Model Predictive Control (MPC) into the RL procedure~\cite{r13}. This work relied on sophisticated and expensive hardware, while its learning efficiency and prediction horizon were also limited by model bias and the computational burden of MPC.
Dropout-based Probabilistic Ensembles with Trajectory Sampling (DPETS) employed probabilistic neural networks to mitigate the model bias in MBRL from a Bayesian perspective~\cite{r14}. 
DPETS shows promising results on MuJoCo benchmarks, yet its implementation challenges for dexterous hand manipulation in terms of model expressiveness and control frequency requirements, remain unaddressed.

To address this issue, Goal-Conditioned Probabilistic Model Predict Control (GC-PMPC) was proposed based on DPETS. To meet the demands for model expressiveness and generalization capabilities in dexterous hand multi-goal manipulation tasks, the proposed method designed a novel architecture of probabilistic neural networks ensembles with a training procedure that considers the predicted uncertainties. An Asynchronous MPC policy was proposed to improve the control frequency while a state smoothing mechanism was incorporated into the policy for superior robustness. The evaluations of GC-PMPC on four Shadow Hand simulation tasks with randomly generated goals demonstrated advantages in both learning efficiency and control performance over related model-free and model-based baselines. We further implemented GC-PMPC on a low-cost dexterous hand DexHand 021 equipped with $12$ active DOFs and $5$ tactile sensors\footnote{More details of DexHand 021 can be found at:~\url{https://www.dex-robot.com/en/productionDexhand}.} to learn a die rotation task with three target poses (with faces 4, 5, and 2 facing upward). Using single-camera for pose detection, our method learned this multi-goal task within 14000 steps interactions, approximately $80$ minutes, demonstrating significant advantages over baselines (Fig.~\ref{figure:1}). The contributions of this paper was summarized as:
\begin{enumerate} 
	\item Developed probabilistic neural networks within the MBRL framework that adapts to dexterous hand multi-goal manipulation tasks, featuring superior generalization and expressive capabilities.
	\item Proposed an asynchronous MPC policy to address the control frequency requirements of dexterous hand systems, complemented by a smoothing mechanism that enhances the robustness of policy in manipulation.
	\item Beyond demonstrating advantages in Shadow Hand simulation scenarios, the proposed method successfully drove a low-cost dexterous hand to learn a die rotation task with multiple goals, achieving superior efficiency and performance compared to baselines.
	\end{enumerate}

The remainder of this paper is organized as follows. Section \ref{S2} provided the problem statement of simulated and real-world dexterous hand system. Section \ref{S3} detailed the proposed method. The experimental results were presented and analyzed in Section~\ref{S4}. The conclusion was given in Sections \ref{S5}.

\begin{figure*}[t]
\centering
\includegraphics[width=1.95\columnwidth]{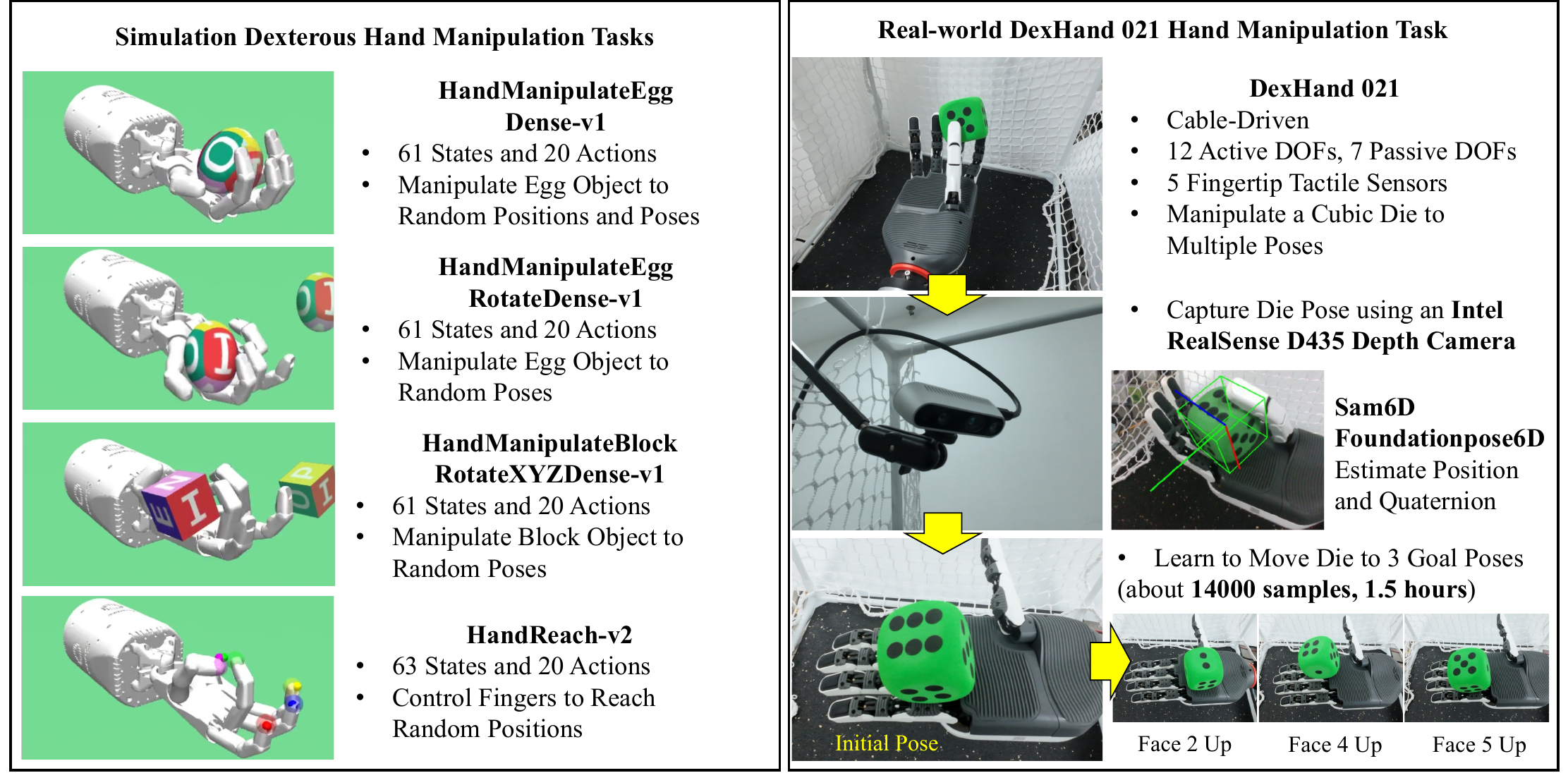}
\caption{Manipulation Scenarios on simulated and real-world dexterous hand systems.}
\label{figure:settings}
\end{figure*}

\section{Problem Statement}\label{S2}
\subsection{Model-based Reinforcement Learning}\label{S2-1}

DPETS~\cite{r14} described the target problem as a Markov decision process (MDP). At time step $t$, the state $\bm{s}_t$ and action $\bm{a}_t$ determined the system's transition to $\bm{s}_{t+1}$ through unknown dynamics $\bm{s}_{t+1} = f(\bm{s}_t, \bm{a}_t)$. A reward $\mathcal{R}(\bm{s}_t, \bm{a}_t)$ was then obtained.
DPETS employed $B$ independent neural networks as ensembles, each network had weights $\bm{W}_b$ with $M$ Dropout particles $\{\bm{z}_m\}_{m=1}^M$ that follow a Bernoulli distribution to capture the uncertainty~\cite{pmlr-v48-gal16}. The ensembles were utilized to model the system dynamics and drive the MPC policy to maximize the cumulative reward with horizon $H$:
\begin{gather} 
\begin{split} 
\bm{s}_{t+1} &= \frac{1}{BM}\sum_{b=1}^{B}\sum_{m=1}^{M}\hat{f}_{\bm{W}_b, \bm{z}_{m}}(\bm{s}_{t}, \bm{a}_{t}),\\
\hat{f}_{\bm{W}_b, \bm{z}_{m}}(\bm{s}_{t}, \bm{a}_{t})&\sim\mathcal{N}\big(\bm{\mu}_{\bm{W}_b, \bm{z}_{m}}(\bm{s}_{t}, \bm{a}_{t}), \bm{\Sigma}_{\bm{W}_b, \bm{z}_{m}}(\bm{s}_{t}, \bm{a}_{t})\big),\\
\left[\bm{a}_t^*, \ldots, \bm{a}_{t+H-1}^*\right] &=\argmax \sum_{h=0}^{H-1}\mathcal{R}\left(\bm{s}_{t+h}, \bm{a}_{t+h}\right)
\label{eq_mpc}
\end{split}
\end{gather}
where $\bm{\mu}_{\bm{W}_b, \bm{z}_{m}}(\cdot)$ and $\bm{\Sigma}_{\bm{W}_b, \bm{z}_{m}}(\cdot)$ were the output of the neural network with weights $\bm{W}_b$ and Dropout particles $\bm{z}_{m}$. The optimal actions were searched by Cross Entropy Method (CEM). A close-loop policy $\pi:\bm{s}_t\rightarrow\bm{a}_t$ was obtained by executing $\bm{a}_t$, move to the next step and repeat the MPC process. All weights of neural networks were iteratively updated by the collected samples $\mathcal{D}$ during the RL process.

\subsection{Dexterous Hand Manipulation Scenarios}\label{S2-2}

The Dexterous Hand Manipulation Scenarios in this work were illustrated in Fig.~\ref{figure:settings}. 
We evaluated the proposed method by four Shadow Dexterous Hand manipulation tasks based on Gymnasium-Robotics\footnote{Details of each simulation task can be found at:~\url{https://robotics.farama.org/envs/shadow_dexterous_hand/.}}. 
In HandManipulateEggDense-v1, the objective was to control the Shadow Hand to manipulate an egg-shaped object to reach randomly generated target poses and positions. 
HandManipulateEggRotateDense-v1 and HandManipulateBlockRotateXYZDense-v1 focused on controlling manipulating egg-shaped and block-shaped objects to achieve randomly generated target orientations, without considering position constraints. 
HandReach-v2 aimed to control the Shadow Hand's fingers to reach randomly generated desired positions. 
The action space for all tasks consisted of joint angles for the Shadow Hand's 20 active DOFs, with a control frequency of $25$ Hz. The observation space encompassed the angles and angular velocities of $24$ joints (with four finger DIP joints excluded from the action space), as well as task-specific information including the target object's position, linear and angular velocities, orientation quaternion, or the 3D positions of five fingertip joints, depending on the task requirements.

For the real-world experiment, we had utilized the cable-driven DexHand 021, developed by DexRobot, to learn the manipulation of a 60mm cubic die from a fixed initial pose to multiple target orientations (with faces 4, 5, and 2 facing upward). The DexHand featured 12 active and 7 passive DOFs, with tactile sensors installed at each fingertip. The action space consists of joint angles for the 12 active DOFs, with a control frequency of $3$Hz. The observation space includes joint angles from the 12 DOFs, normal forces from the 5 fingertip tactile sensors, and the die's 3D position and quaternion pose.
Unlike simulation environments where target object poses and positions were directly accessible, an Intel RealSense D435 depth camera was used to detect the die pose in this scenario. 
Sam6D~\cite{lin2024sam} was utilized for image segmentation based on the die's 3D model and depth camera signals.
Foundationpose6D~\cite{10655554} was then applied to obtain the 3D position and quaternion pose of the manipulated die\footnote{The code of Sam6D and Foundationpose6D is available at \url{https://github.com/JiehongLin/SAM-6D} and \url{https://github.com/NVlabs/FoundationPose}.}. The reward function was defined as:
\begin{gather} 
\begin{split} 
\mathcal{R}(\bm{s}_t, \bm{a}_t)=-\left(  0.1 \cdot \| \bm{p}_t - \bm{p}_t^\star \|_2 + 2\cdot\arccos\left( \left| \langle \bm{q}_t, \bm{q}_t^\star \rangle \right| \right) \right)
\label{eq_reward}
\end{split}
\end{gather}
where $\bm{p}_t$ and $\bm{p}_t^\star$ were the current and target positions, and $\bm{q}_t$ and $\bm{q}_t^\star$ were the current and target quaternion poses. The target information was derived from the die's position and pose when the desired face was oriented upward. Given the DexHand 021's structural limitations including lack of wrist joints, high palm friction, and difficulty in fine position adjustments, we assigned a small weight to the position error term and primarily focused on the pose of die. Unlike simulation tasks with completely random generated goals, we trained DexHand 021 sequentially on goals requiring faces 2, 4, and 5 to face upward. After mastering each task, we continued the training process by directly modifying the reward function without interruption.

\section{Approach}\label{S3}
\subsection{Goal-Conditioned Neural Network Ensembles}\label{S3-1}
In multi-goal dexterous hand manipulation tasks, samples from the high-dimensional state-action space usually show non-uniform distributions. It led to several challenges for the original probabilistic neural network ensembles in DPETS including slow convergence, reduced exploration efficiency, and insufficient prediction accuracy. To address these issues, GC-PMPC incorporated Batch Normalization into the probabilistic neural network ensembles, building on its proven ability to significantly enhance the training efficiency and stability of neural networks~\cite{santurkar2018does}. This modification effectively reduces the negative impact of large data scale differences in multi-objective tasks and high-dimensional systems on learning speed. The improved prediction in GC-PMPC satisfied:
\begin{gather} 
\begin{split} 
\bm{s}_{t+1} = \frac{1}{BM}&\sum_{b=1}^{B}\sum_{m=1}^{M}\hat{f}_{\bm{W}_b, \bm{z}_{m}}\big(\text{BN}(\bm{s}_{t}), \text{BN}(\bm{a}_{t})\big),\\
\text{BN}(\bm{s}_t) &= \frac{\bm{s}_t - \bm{\mu}_{\bm{s}}}{\bm{\sigma}_{\bm{s}}}, \quad \text{BN}(\bm{a}_t) = \frac{\bm{a}_t - \bm{\mu}_{\bm{a}}}{\bm{\sigma}_{\bm{a}}}
\label{eq_bn}
\end{split}
\end{gather}
where $\bm{\mu}_{\bm{s}}, \bm{\sigma}_{\bm{s}}, \bm{\mu}_{\bm{a}}, \bm{\sigma}_{\bm{a}}$ were the mean and standard deviation of all states and actions collected in the current samples set $\mathcal{D}$. They were recalculated iteratively before each model update.

\begin{figure*}[t]
\centering
\includegraphics[width=1.8\columnwidth]{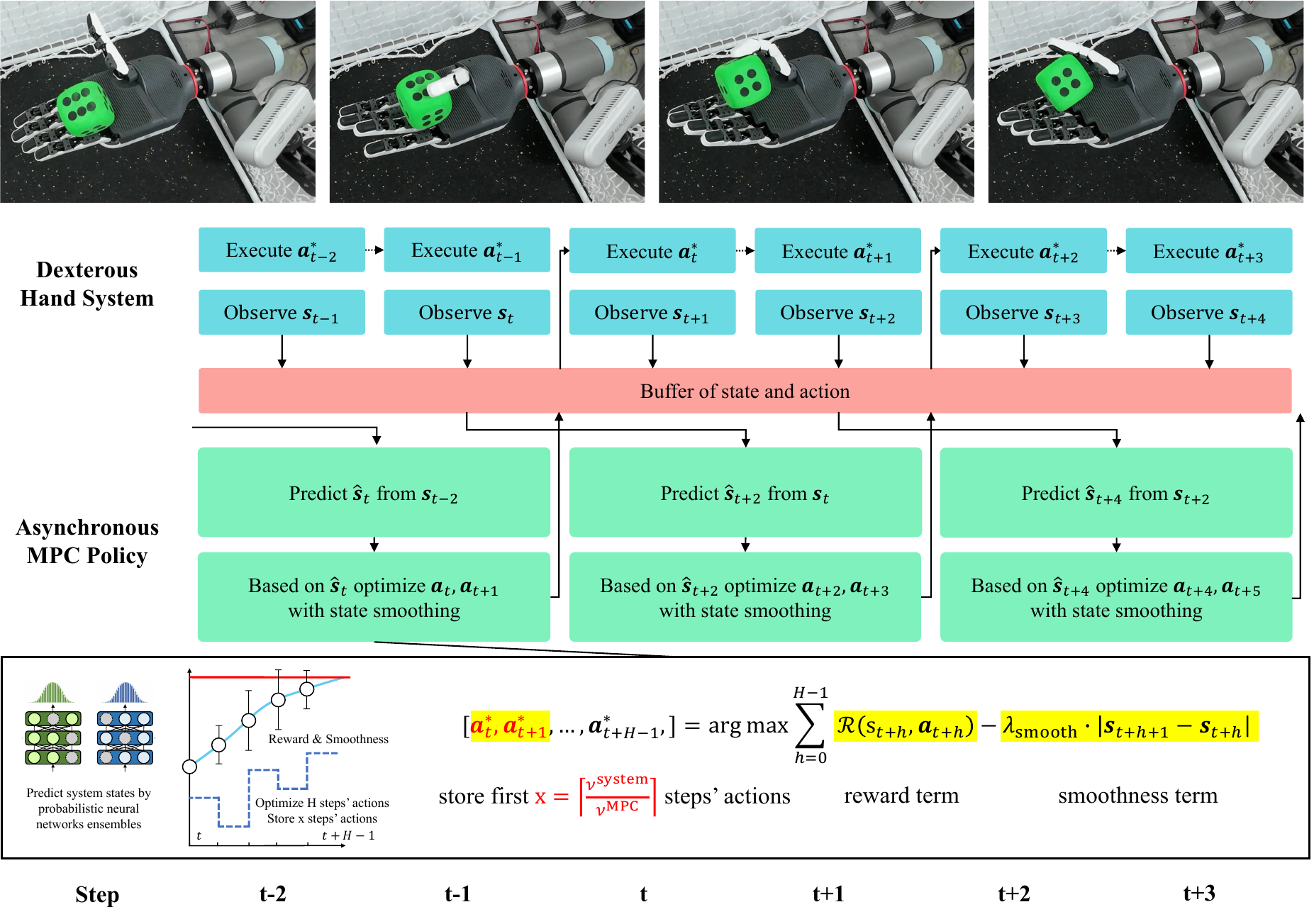}
\caption{Principles of the asynchronous MPC policy ($x=2$) designed to match the control frequency of real-world dexterous hand systems.}
\label{figure:ampc}
\end{figure*}

We further expanded upon DPETS's loss function, which considers two-step model prediction errors, by adding an additional prediction variance constraint term. By limiting the variance of predictions from the probabilistic neural network ensembles, GC-PMPC improved the efficiency and stability in model training, providing a more reliable probabilistic model for the MPC policy specialized for dexterous hand. Define $N$ samples of two continuous steps states and actions as $\{\bm{s}_t^n, \bm{a}_t^n, \bm{s}_{t+1}^n, \bm{a}_{t+1}^n, \bm{s}_{t+2}^n\}_{n=1}^{N}$, the loss function of the $b$-th neural network in ensembles followed:
\begin{gather} 
\begin{split}
&\mathcal{L}_b = \frac{1}{N} \sum_{n=1}^{N}\bigg[
\underbrace{
        \bm{E}_{\bm{W}_b}^{\top} \bm{\Sigma}_{\bm{W}_b}^{-1}(\bm{s}_{t}^n, \bm{a}_{t}^n)\bm{E}_{\bm{W}_b} + \log \det \bm{\Sigma}_{\bm{W}_b}(\bm{s}_{t}^n, \bm{a}_{t}^n)
}_{\text{First step loss}}
\\&+
\underbrace{
        \bm{E}_{\bm{W}_b}^{\prime\top} \bm{\Sigma}_{\bm{W}_b}^{-1}(\bm{s}_{t+1}^n, \bm{a}_{t+1}^n)\bm{E}_{\bm{W}_b}^{\prime} + \log \det \bm{\Sigma}_{\bm{W}_b}(\bm{s}_{t+1}^n, \bm{a}_{t+1}^n)
}_{\text{Second step loss}
}
\\&+
\underbrace{\Delta\cdot\big(\bm{\Sigma}_{\bm{W}_b}(\bm{s}_{t}^n, \bm{a}_{t}^n)+\bm{\Sigma}_{\bm{W}_b}(\bm{s}_{t+1}^n, \bm{a}_{t+1}^n)\big)}_{\text{Penalty for over-large predicted variances}}\bigg] \!\!+\!\! \sum_{l=1}^L\lambda_l\cdot \|\mathbf{W}_b^l\|_2^2.
\label{eq_RMC_loss}
\end{split}
\end{gather}
The first two terms were the predicted errors for the first and second steps where $\bm{E}_{\bm{W}_b}=\left[\bm{\mu}_{\bm{W}_b}\left(\bm{s}_{t}^n,\bm{a}_{t}^n\right)-\bm{s}_{t+1}^n\right], \bm{E}_{\bm{W}_b}^{\prime}=\left[\bm{\mu}_{\bm{W}_b}\left(\bm{s}_{t+1}^n,\bm{a}_{t+1}^n\right)-\bm{s}_{t+2}^n\right]$ calculated the difference between the output mean and the ground truth. The third term penalized excessive prediction variance controlled by parameter $\Delta$ which was set to $30$ for all experiments. The final term penalized oversized weights and biases in $\bm{W}_b$ across the $L$-layer neural network. 
During training, the Dropout particles hidden in Eq.~\eqref{eq_RMC_loss} were randomly selected from another independent set to encourage better model generalization.

\begin{figure*}
\centering
\includegraphics[width=1.95\columnwidth]{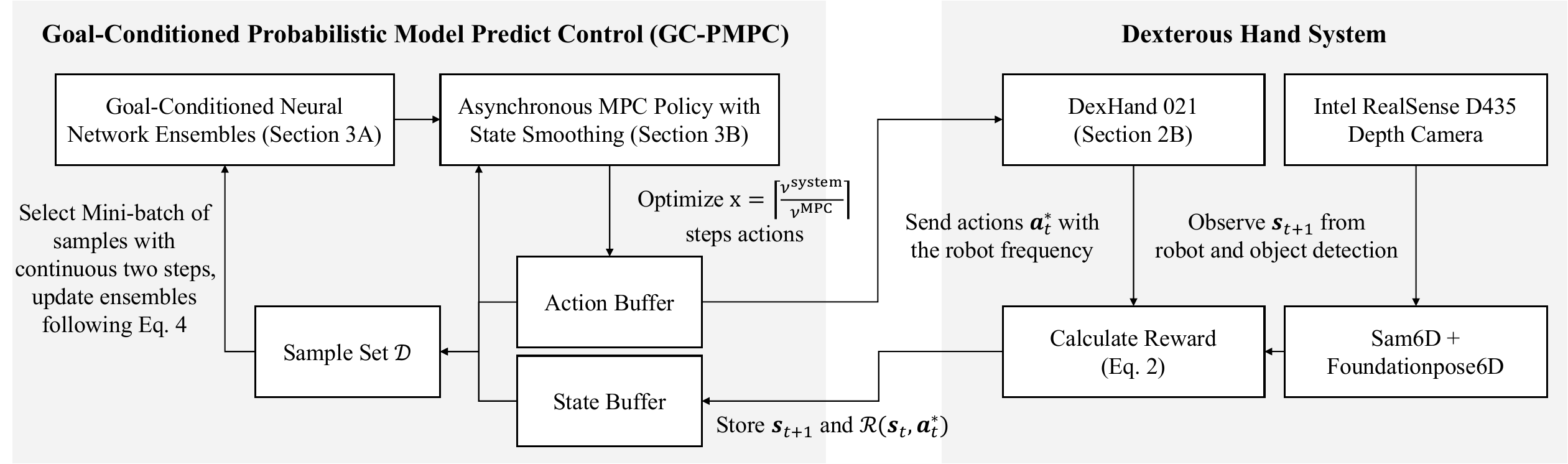}
\caption{Workflow of GC-PMPC on the real-world DexHand 021 System.}
\label{figure:workflow}
\end{figure*}

\subsection{Asynchronous MPC Policy with State Smoothing}\label{S3-2}

Due to the high computational demands of probabilistic neural network ensembles in MPC which typically cannot match the control frequency needed for real dexterous hand systems, GC-PMPC proposed a novel asynchronous mechanism to bridge such a frequency gap. This is accomplished by executing MPC optimization in a separate asynchronous thread and storing multiple action steps, effectively decoupling it from the real-world dexterous hand thread. Define the control frequency of the target system as $\nu^{\text{system}}$, the control frequency of the MPC policy as $\nu^{\text{MPC}}$, the predict and output step was calculated as $x=\lceil \nu^{\text{system}} / \nu^{\text{MPC}} \rceil$. 

\begin{figure*}
\centering
\includegraphics[width=1.95\columnwidth]{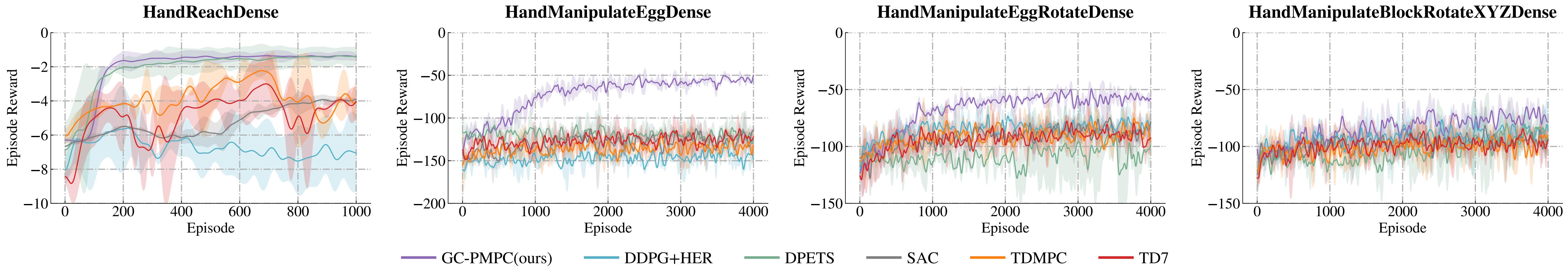}
\caption{Learning curves of GC-PMPC and other baselines in simulated manipulation tasks. The shaded region represents the corresponding standard deviation.}
\label{figure:simulated_curves}
\end{figure*}

\begin{figure*}
\centering
\includegraphics[width=1.95\columnwidth]{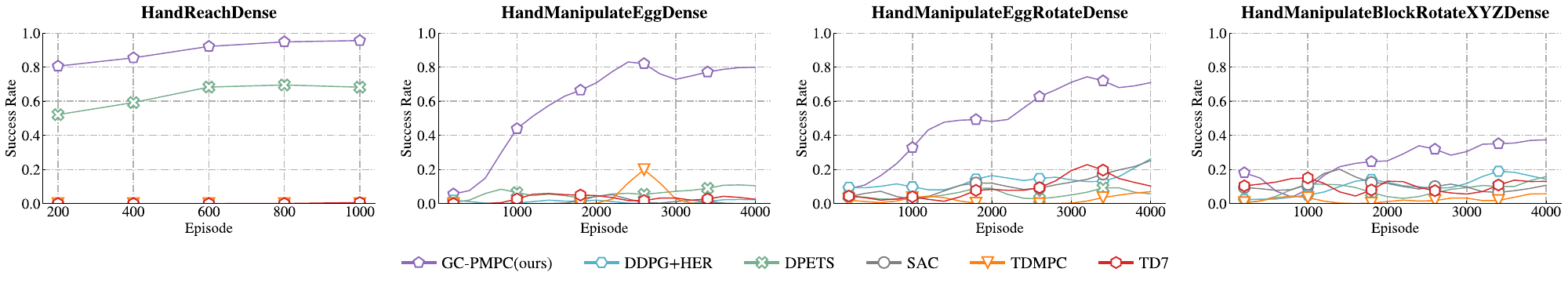}
\caption{Success rates of GC-PMPC and other baselines in simulated manipulation tasks. The rates were summarized by $20$ random generated goals.}
\label{figure:simulated_ratios}
\end{figure*}

Figure~\ref{figure:ampc} illustrated the principle of the proposed asynchronous MPC policy with $x=2$. 
At time step $t$, the dexterous hand system executed $x$ the actions $\bm{a}_{t}, \bm{a}_{t+1}^*$ pre-planned on an independent thread from the action buffer and sequentially observed the resulting states $\bm{s}_{t+1}, \bm{s}_{t+2}$. Concurrently, the MPC policy thread retrieved the previous state $\bm{s}_{t}$ from the buffer and estimated the future state $\hat{\bm{s}}_{t+2}$ based on $\bm{a}_{t}^{*}, \bm{a}_{t+1}^{*}$. The MPC policy searched the optimal actions based on $\hat{\bm{s}}_{t+2}$ following Eq.~\eqref{eq_mpc}. The first $x$ actions ($\bm{a}_{t+2}^{*}, \bm{a}_{t+3}^{*}$) were stored to the buffer and then sent to the dexterous hand. When $t=0$, $\bm{s}_{0}$ was observed by executing zero actions in advance.

The asynchronous MPC policy inevitably led to skipping the actual system states and relying on predicted states, which made GC-PMPC more sensitive to model errors. To address this issue and optimize the balance between exploitation and exploration in the policy, a state smoothing term was added to the MPC optimization objective to reduce sudden state changes that could be triggered by over-large control actions:
\begin{gather} 
\begin{split} 
\argmax_{\bm{a}_t^*, \ldots, \bm{a}_{t+H-1}^*} \sum_{h=0}^{H-1}\mathcal{R}\left(\bm{s}_{t+h}, \bm{a}_{t+h}\right)+\lambda_{\text{smooth}}\cdot|\bm{s}_{t+h+1}-\bm{s}_{t+h}|
\label{eq_ampc}
\end{split}
\end{gather}
where the smoothness parameter was set as $\lambda_{\text{smooth}}=0.01$ for the real-world experiment.

\subsection{Overview of GC-PMPC}\label{S3-3}

Fig.~\ref{figure:workflow} illustrates the workflow of GC-PMPC on the real-world DexHand 021. The model knowledge is provided by a probabilistic neural network ensemble (Section~\ref{S3-1}) for the Asynchronous MPC policy (Section~\ref{S3-2}). At each time step, GC-PMPC optimizes control actions based on system frequency requirements. The hand's state, including joint positions, tactile feedback, and dice pose (from Sam6D and Foundationpose6D), is used to compute rewards (Eq.~\eqref{eq_reward}) and stored in the state buffer. The neural network ensemble is updated after each episode using Eq.~\eqref{eq_RMC_loss} with randomly sampled consecutive state-action pairs from $\mathcal{D}$.

\begin{figure*}[t]
\centering
\includegraphics[width=2.0\columnwidth]{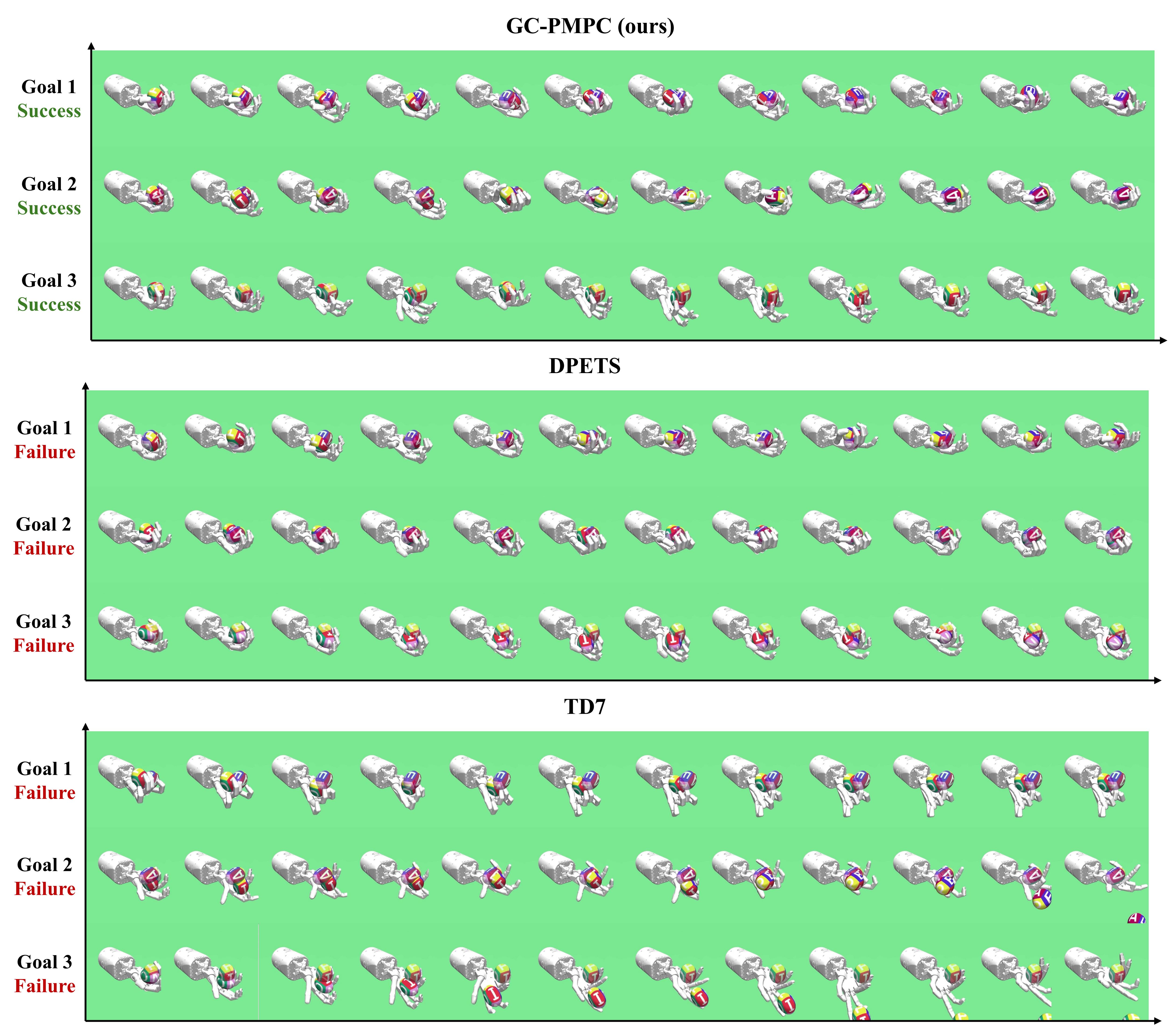}
\caption{Testing trajectories of learned GC-PMPC, DPETS and TD7 agents in HandManipulationEggDense task with three goals.}
\label{figure:simulated_case}
\end{figure*}

\section{Experimental Results}\label{S4}
\subsection{Experimental Settings}\label{S4-1}

For all experiments, GC-PMPC employed an ensemble of $B=5$ probabilistic neural networks, where each network was structured as a multilayer perceptron with three hidden layers $[512, 512, 512]$. The MPC policy horizon was set to $H=50$ steps for simulation experiments and $H=30$ steps for real-world implementations.
For the compared methods, we selected Soft Actor Critic (SAC) and Deep Deterministic Policy Gradient (DDPG) with Hindsight Experience Replay (HER) as traditional model-free RL baselines, and TD7~\cite{fujimoto2023sale} as a state-of-the-art model-free algorithm. 
For MBRL, we chose Probabilistic Ensembles with Trajectory Sampling (PETS)~\cite{chua2018deep} as the traditional baseline, while the original DPETS~\cite{r14} and Temporal Difference MPC  (TDMPC)~\cite{hansen2022temporal} served as state-of-the-art benchmarks.
Each episode was defined as $50$ steps. For simulation tasks, we initialized training with $5000$ samples generated by random actions, while the real-world experiment used $4000$ samples collected by a data glove for warm-up.

\begin{figure*}
\centering
\includegraphics[width=1.8\columnwidth]{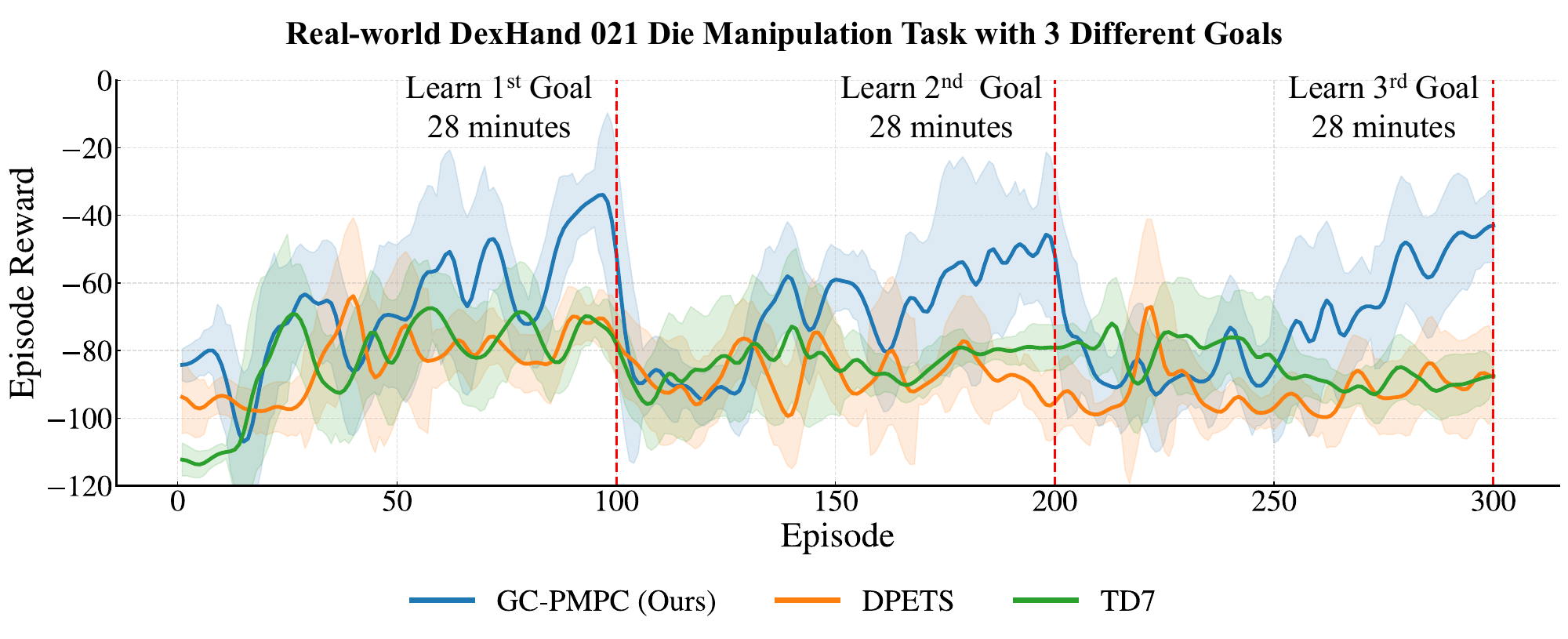}
\caption{Learning curves of GC-PMPC, DPETS and TD7 in the real-world experiment. The shaded region represents the corresponding standard deviation.}
\label{figure:real_curves}
\end{figure*}

The simulation experiments were conducted through three independent trials on a high-performance computing server equipped with an AMD EPYC 7763 64-Core CPU, eight NVIDIA GeForce RTX 4090 GPUs, 512GB memory, and Ubuntu 20.04 OS.
For the real-world experiments, we utilized three separate workstations: (1) GC-PMPC ran on a laptop with Intel i9-14900HX CPU, NVIDIA RTX 4090 Laptop GPU, and 64GB memory; (2) DexHand's control system was operated on a workstation with Intel® Xeon(R) W-2295 CPU, NVIDIA GeForce RTX 3090 GPU, and 64GB memory; (3) Object detection was performed on a workstation with Intel i9-13900K CPU, NVIDIA GeForce RTX 4080 GPU, and 64GB memory. All systems ran Ubuntu 20.04 OS.

\subsection{Simulation Results}\label{S4-2}
The learning curves of all compared approaches' episode rewards over four simulated manipulation scenarios were demonstrated in Fig.~\ref{figure:simulated_curves}.
Specific success thresholds were defined to evaluate task completion. In the position reaching task, success occurred when finger position errors fell below $2$mm at the end of episode. For manipulation tasks, success required orientation errors under $0.5$ radians and position errors under $5$mm. We evaluated each algorithm through $20$ trials with randomized goals, with results shown in Fig.~\ref{figure:simulated_ratios}.

For the relatively simple Hand Reach task, GC-PMPC demonstrated remarkable efficiency by mastering the task in $200$ episodes ($400$ seconds), significantly outperforming other baselines in terms of learning speed and exhibiting smaller reward variance (indicating superior policy stability). Regarding success rates, the proposed method consistently outperformed the original DPETS. Among model-free RL approaches, none of model-free RL baselines met the success requirements for this task. Based on our empirical observations, model-free approaches eventually converged to performance levels comparable to GC-PMPC after approximately $500000$ steps, which clearly demonstrates the superior learning efficiency of our proposed method.
In the Egg Manipulation task with both random goals of positions and poses, GC-PMPC demonstrated overwhelming advantages. After $100$ episodes ($30$ minutes) of learning, it outperformed all other baselines, achieving a success rate of $80\%$ while other approaches showed virtually no signs of learning.

In the Manipulation tasks focused on reaching random poses, removing position constraints for pose achievement led to sparser rewards and increased difficulty. For the egg manipulation tasks, GC-PMPC demonstrated superior performance with smoother reward curves and higher upper limits, achieving approximately $70\%$ success rate after one hour's learning, while other baselines reached only $20\%$. Notably, without the improved goal-conditioned neural network ensembles and state smoothing in MPC policy, the original DPETS showed significant reward variance, revealing its limitations in learning multi-goal tasks. The Block Manipulation task proved more challenging as it required coordinated five-finger actions rather than primarily wrist rotation. In this task, GC-PMPC maintained its advantages in learning efficiency and reward ceiling, achieving more than double the success rate of other baselines after one hour's learning.

\begin{figure}
\centering
\includegraphics[width=0.95\columnwidth]{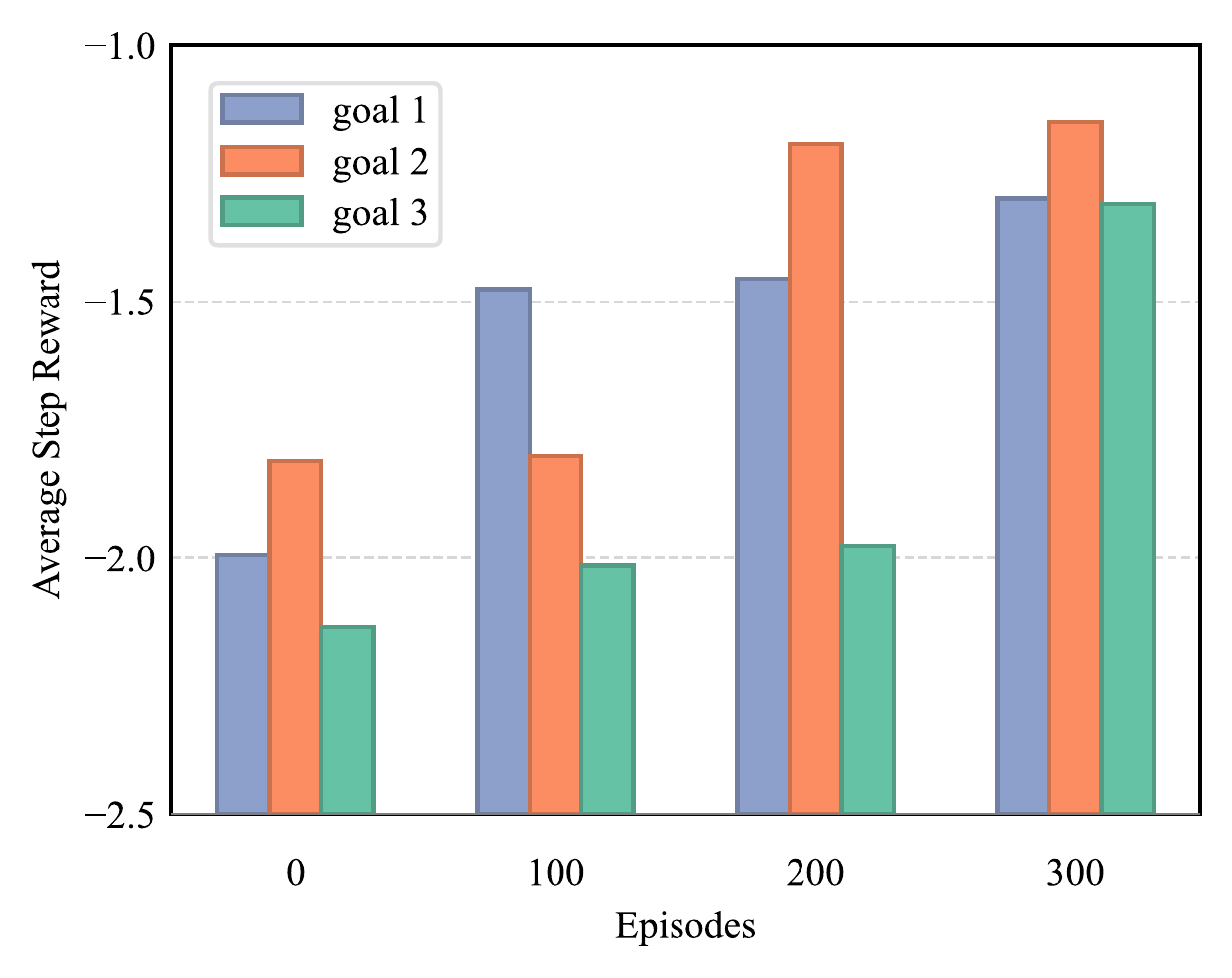}
\caption{Testing rewards of GC-PMPC for three goals during learning processes with different goal intervals.}
\label{figure:real_bar}
\end{figure}

As a case study, Fig.~\ref{figure:simulated_case} presented the control trajectories of GC-PMPC, DPETS, and TD7 in the HandManipulationEggDense task with three randomly generated goals. The proposed GC-PMPC successfully achieved all goals by manipulating the egg-shaped object to desired positions and poses. However, DPETS exhibited insufficient control capabilities with notable position errors, mainly due to its limited model expressiveness in capturing complex hand dynamics. TD7 performed worst, with its Actor networks generating unnatural finger motions that led to substantial errors and object dropping. These results are consistent with the performance comparisons shown in Figs.~\ref{figure:simulated_curves} and~\ref{figure:simulated_ratios}.

\begin{figure*}[t]
\centering
\includegraphics[width=1.8\columnwidth]{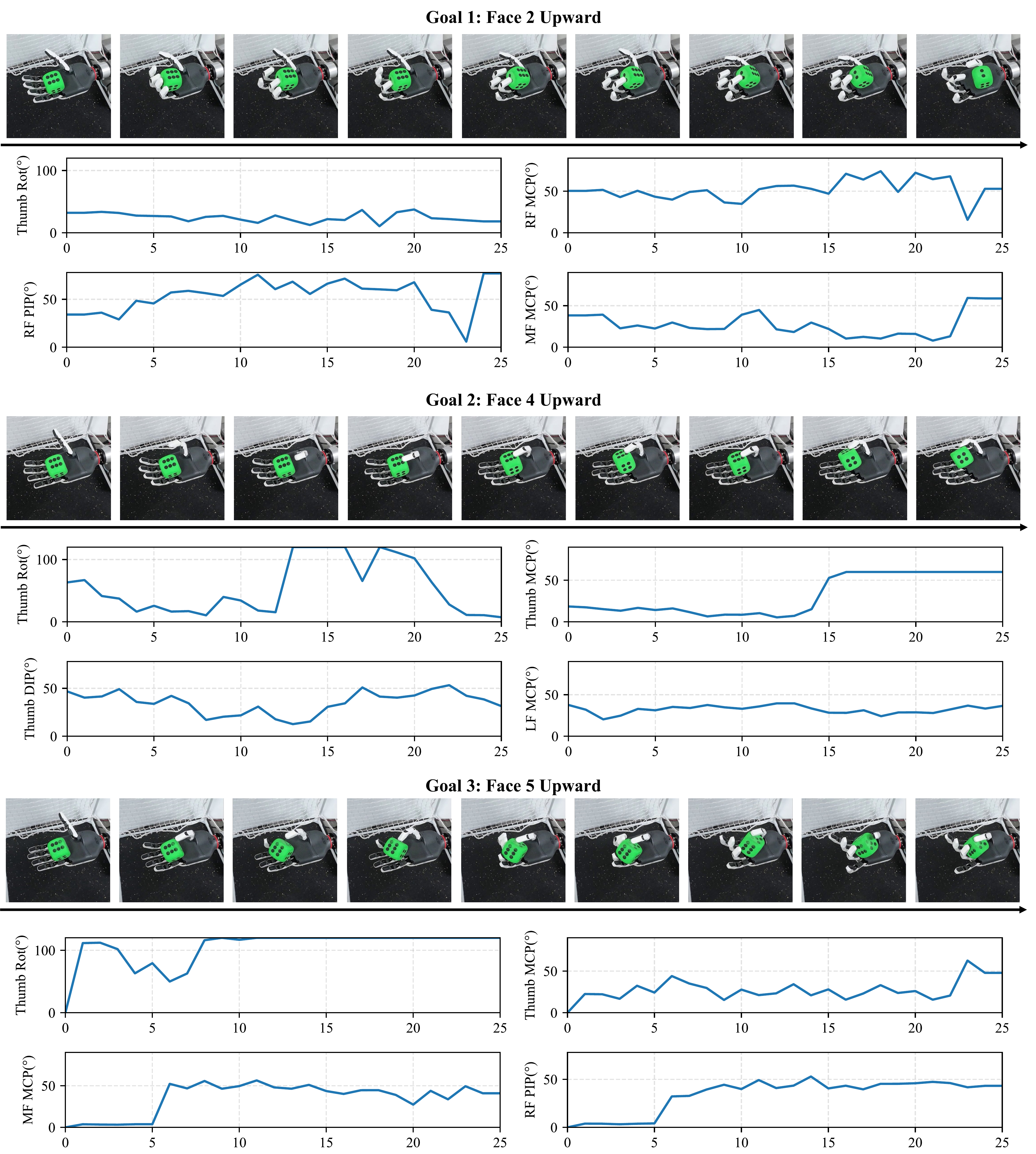}
\caption{Testing trajectories of learned GC-PMPC agents in controlling DexHand 021 to manipulate the die to reach three goal poses.}
\label{figure:real_case}
\end{figure*}

\subsection{Real-world Hardware Results}\label{S4-3}
In the real-world hardware experiment, we sequentially trained the DexHand 021 to manipulate the die to reach three distinct goal positions (2, 4, and 5 dots facing upward), with each goal comprising $100$ episodes, $5000$ steps, approximately $28$ minutes of training time. We employed DPETS and TD7 as model-based and model-free baselines, respectively. Figure~\ref{figure:real_curves} presented a comparative analysis of episode rewards throughout the training process, following Eq.~\eqref{eq_reward}, with an additional penalty of $-2$ assigned when the die was dropped out of hand.
The results demonstrated that GC-PMPC was the only method exhibiting stable learning progression, achieving significant reward improvements within each $28$-minute goal interval. In contrast, other methods showed negligible improvement in control behavior throughout the entire $80$-minute learning process. Furthermore, Figure~\ref{figure:real_bar} illustrated the comparative test results across all three goals during different learning phases. Notably, due to the model-based nature and the flexibility of reward function adaptation in the MPC policy, GC-PMPC maintained performance on previously learned goals while acquiring new ones.
The proposed successfully learned control policies for manipulating the die to achieve three different goal poses within approximately $80$ minutes' exploration, demonstrating remarkable advantages in both learning efficiency and policy stability for multi-task scenarios.

Figure.~\ref{figure:real_case} illustrated the successful manipulation trajectories of GC-PMPC in achieving three distinct goal poses, along with the corresponding trajectories of the four active DOFs over $25$ time steps. For the first goal, GC-PMPC leveraged the coordinated movement of two ring finger DOFs (RF PIP and RF MCP) to generate three successive die-pushing actions toward the wrist at steps $10$, $15$, and $20$, achieving the desired die orientation. The less active thumb and middle finger DOFs (Thumb Rot and MF MCP) primarily served to prevent the die from falling. For the second goal, GC-PMPC primarily coordinated three thumb DOFs (Thumb Rot, Thumb MCP, and Thumb DIP), with Thumb Rot leading two movements to rotate the die laterally until the four-dot face upward. In the most challenging task of maintaining the die's five-dot face upward, GC-PMPC first utilized two thumb DOFs (Thumb Rot, Thumb MCP) to position the die at an appropriate angle, then completed the manipulation through coordinated actions of the middle and ring fingers (MF MCP and RF PIP).

\subsection{Engineering Challenges in Real-World Dexterous Hand}\label{S4-4}

Despite promising simulation results, the real-world implementation of GC-PMPC revealed several challenges.
Environmental noise and hardware limitations significantly impacted the performance in terms of both learning efficiency and control behavior. 
The DexHand 021 exhibited inter-finger interference, causing motor stalling and suboptimal action execution. The system also demonstrated temperature-dependent variations in both action execution and state feedback, which were absent in simulation.
The proprioceptive capabilities of the real-world dexterous hand were also limited compared to the simulated Shadow Hand, with only fingertip tactile sensors and no direct joint velocity measurements.
Most significantly, the 12-DOF configuration showed reduced learning efficiency compared to the 20-DOF simulated Shadow Hand. The absence of wrist articulation proved particularly limiting, as simulation studies demonstrated its crucial role in object manipulation. The die manipulation task, relying solely on finger coordination, required greater lateral freedom of fingers. However, the DexHand 021 could only achieve synchronized extension and flexion of four fingers (excluding the thumb) and therefore had degraded control performance.

\section{Conclusions}\label{S5}
In this paper, we present Goal-Conditioned Probabilistic Model Predictive Control (GC-PMPC), a novel MBRL approach for multi-goal dexterous hand manipulation. Our proposed approach incorporates probabilistic neural network ensembles enhanced with Batch Normalization and variance prediction penalties, thereby improving model expressiveness and generalization capabilities in high-dimensional state-action spaces. Furthermore, an asynchronousMPC policy with state smoothing is introduced to address the challenges of control frequency and policy robustness in practical dexterous hand systems.
Comprehensive experimental evaluations conducted on four Shadow Hand simulation manipulation scenarios demonstrate that GC-PMPC significantly outperforms state-of-the-art model-free methods (DDPG+HER, SAC, TD7) and model-based approaches (PETS, DPETS, TDMPC) in terms of both convergence speed and success rate. Real-world implementation on the DexHand 021 system, equipped with 12 active DOFs and 5 tactile sensors, successfully accomplished a multi-goal die manipulation task within approximately 80 minutes' learning using  single-camera pose detection. These results not only validate the superior efficiency and adaptability of the proposed method on a cost-effective dexterous hand platform but also demonstrate the significant potential of probabilistic MBRL with MPC policies for learning challenging dexterous manipulation tasks.

\section*{Acknowledgements}
We extend our sincere gratitude to the DexRobot engineering team for their invaluable support during our real-world experiments. Die manipulation tasks pose significant challenges for cost-effective dexterous hands, and their improvements to hardware thermal management and joint structural design were crucial for the successful completion of our experiments.

\bibliographystyle{ieeetr}
\bibliography{paper}

\end{document}